\documentclass[letterpaper, 10 pt, conference]{ieeeconf}  

\IEEEoverridecommandlockouts                              

\usepackage{graphics} 
\usepackage{amsmath} 
\usepackage{amssymb}  

\usepackage{hyperref}       
\usepackage{url}            
\usepackage{booktabs}       
\usepackage{amsfonts}       
\usepackage{nicefrac}       
\usepackage{microtype}      
\usepackage{dsfont}
\usepackage{algorithm2e}
\usepackage{graphicx} 
\usepackage{xcolor}	
\usepackage{cite}
\usepackage{subcaption}
\title{\LARGE \bf
Information-theoretic Model Identification and Policy Search using Physics Engines with Application to Robotic Manipulation
}

\author{Shaojun Zhu$^{1}$, Andrew Kimmel$^{1}$, Abdeslam Boularias$^{1}$
\thanks{$^{1}$Shaojun Zhu, Andrew Kimmel, and Abdeslam Boularias are with the Department of Computer Science, Rutgers University, New Jersey, USA {\tt\small{ \{shaojun.zhu, andrew.kimmel, abdeslam.boularias\}@cs.rutgers.edu}}}%
}

\begin{document}

\maketitle
\thispagestyle{empty}
\pagestyle{empty}

\begin{abstract}

We consider the problem of a robot learning the mechanical properties of objects through physical interaction with the object, and introduce a practical, data-efficient approach for identifying the motion models of these objects.  The proposed method utilizes a physics engine, where the robot seeks to identify the inertial and friction parameters of the object by simulating its motion under different values of the parameters and identifying those that result in a simulation which matches the observed real motions. The problem is solved in a Bayesian optimization framework. The same framework is used for both  identifying the model of an object online and searching for a policy that would minimize a given cost function according to the identified model. Experimental results both in simulation and using a real robot indicate that the proposed method outperforms  state-of-the-art model-free reinforcement learning approaches. 


\end{abstract}

\section{Introduction}

\begin{figure}
\centering
\includegraphics[width=0.49\textwidth]{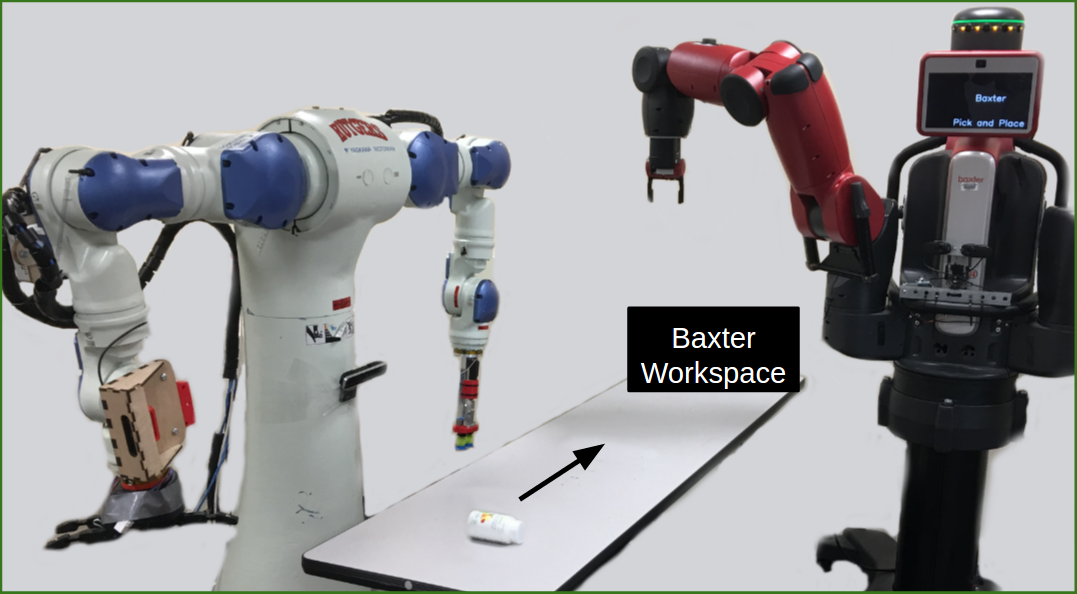}
\caption{Proposed Experiment: The cylindrical object (bottle) on the table is unknown,  and the Baxter robot needs to grasp it but it cannot reach it. The Motoman SDA10F robot on the left can reach the object. The Motoman pushes with the object a few times, identifies its mechanical properties, and attempts to roll it safely into Baxter's reachable workspace, without dropping it off the table.}
\label{fig:ex}
\end{figure}

Consider the scenario shown in Figure \ref{fig:ex}, where a robot (Motoman) assists another robot (Baxter) that cannot reach its desired object.  Due to the placements of the robots in the scene, the intersection of each robot's reachable workspace is empty, which restricts the robots from executing a ``direct hand-off'' maneuver.  In this case, the Motoman robot must exploit the dynamic physical properties of the object in order to ``slide'' it over to the Baxter robot.  Ideally, this action would happen without intervention or assistance from an outside operative, such as a human.

Learning the physical properties of an object and predicting its motion under physical interaction is a critical aspect of this challenge. If the robot simply executes a maximum velocity push on the object, the result could cause the object to leave the robot's workspace (i.e. falling off the table), which is undesirable as it would ruin the autonomous behavior of the system. 



This paper proposes a data-efficient approach for motion prediction by utilizing a physics engine and learning the physical parameters through black-box Bayesian optimization. Specifically, the objective of the method is to predict the motion of an object when acted upon by a robotic hand. First, a real robot is used to perform some random pushing action  with an object on a tabletop~\cite{agrawal2016learning}. Both the initial and final configurations of the object and the hand are recorded. Instead of learning the object's motion explicitly, a Bayesian optimization technique is used to identify relevant physical parameters, such as mass and friction, through the physics engine simulation. To predict the motion of the object under a new action, the learned parameters can be used to simulate the action in a physics engine. The results of this simulation can then be used by the robot to predict the effect of its action on the object. To solve the challenge in Figure \ref{fig:ex}, the same Bayesian optimization technique is used to search the optimal control policy for the robotic hand pushing the object. 

\section{Related work}

Several physics engines have been used for simulating dynamics of robots as well as the objects they interact with. Examples of popular physics engines frequently used in robotics include {\it Bullet}~\cite{Bullet}, {\it MuJoCo}~\cite{MuJoCo}, {\it DART}~\cite{DART}, {\it PhysX}~\cite{PhysX}, {\it Havok}~\cite{Havok}, {\it ODE}~\cite{ODE}, and {\it GraspIt!}~\cite{GraspItSimulator}. A survey and a comparison of these tools are given in~\cite{ErezTT15}.

Data-driven system identification is a popular approach that is at the
core of learning for control techniques. Examples of these techniques
include model-based reinforcement learning for
instance~\cite{Sutton:1998:IRL:551283}. We focus here on works related
to learning mechanical models of unknown objects.  Several cognitive
models that combine Bayesian inference with approximate knowledge of
Newtonian physics have been proposed
recently~\cite{Hamrick2016Cognitive, Chang2016,BattagliaNIPS2016}.
These methods learn probabilistic models from noisy physical
simulations. Nevertheless, these models are built to explain the
learning of Newtonian physics in humans, rather than to be used for
robotic manipulation, which typically requires a higher precision as well as
faster learning and inference times. 



Two high-level approaches exist for solving physical interaction problems, which reside at two extremes of a spectrum. Model-based approaches~\cite{Dogar_2012_7076,LunchMason1996,Mericli2014,isbell:physics:2014, ZhouPBM16} rely on accurate models for
objects and their properties. They are used within standard simulation, planning, and actuation control loops.  A physics-based simulation was used in~\cite{Dogar_2012_7076} for predicting the effects of pushing actions, but the authors considered only flat, well-separated objects on a smooth surface. A nonparametric approach was used in~\cite{Mericli2014} for learning the outcome of pushing large objects (furniture). A Markov Decision Process (MDP) is used in~\cite{isbell:physics:2014} for modeling interactions between objects, however, only simulation results on pushing were reported in that work. Nevertheless, it is prohibitive to manually define perfect and accurate models that
express all types of interactions a robot can experience in the real
world. Other Bayesian model-based techniques, such as PILCO~\cite{Deisenroth:2011fu}, have been proven efficient in utilizing a small amount of data for learning dynamical models and optimal policies. These techniques learn dynamical equations from scratch, unlike our method which assumes that the motion equations are known and provided by a physics engine, and instead concentrates on identifying only the inertial and friction properties of the objects.

Another alternative, which is becoming increasingly popular, addresses these challenges through end-to-end
learning~\cite{agrawal2016learning,EslamiHWTKH16,
FragkiadakiALM15, ullman:cogsci2014,WuYLFT15,ByravanF16,
finn2016deep,ZhangWZFT16,li16arxiv,LererGF16,
DBLP:journals/corr/PintoGHPG16,
DBLP:journals/corr/0003LF16,citeulike:14184576}. This involves the demonstration of many successful examples of physical interaction and learning the controls for solving a problem as a function of the sensing input. These approaches, however, usually require many physical experiments to effectively learn.
The proposed method aims to be more data-efficient, and can quickly adapt online to minor changes in object dynamics. Furthermore, it is not clear for existing methods how uncertainty, a consequence of learning from a small number of data points, could be handled in a
principled way. Note that there is a significant body of work on learning sliding models of objects using {\it white-box} optimization~\cite{ZhouPBM16,
LunchMason1996,DBLP:conf/iros/YuLR15}.  It is not clear, at the moment, if these methods would perform better than the proposed approach. A drawback of white-box methods is that they are often used only in simple setups, such as pushing planar objects~\cite{ZhouPBM16}.

\section{Proposed Approach}
\subsection{System Overview}
To solve the problem of modeling mechanical properties of objects, this paper proposes an online learning approach to identify mass and sliding models of objects using Bayesian optimization.  The goal is to allow the robot to use predefined models of objects, in the form of prior distributions, and to improve the accuracy of these models on the fly by interacting with the objects. This learning process must happen in real time since it takes place simultaneously with the physical interaction. 

\begin{figure*}[t]
\begin{center}
	       \includegraphics[width=\textwidth]{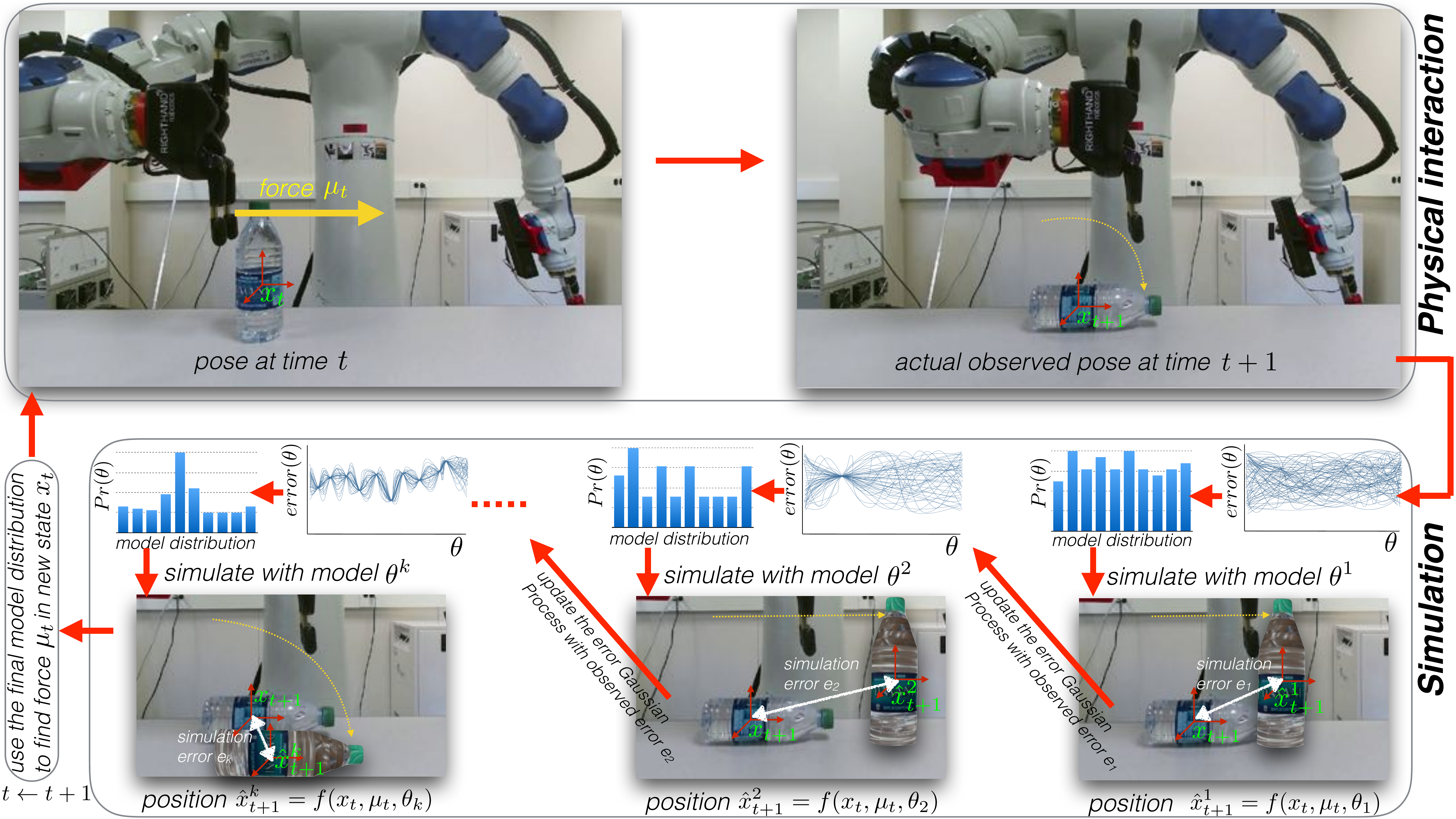}
\end{center}
\caption{Overview of the proposed approach for learning mechanical models of objects with a physics engine.}
\label{systemIDWorkFlow}
\end{figure*}

Figure~\ref{systemIDWorkFlow} shows an overview of the proposed approach. The first step consists of using a pre-trained object detector to detect the different objects present in the scene and estimate their poses by mapping them to a knowledge base of pre-existing 3D mesh models. The proposed method augments the 3D mesh models of each object with the mechanical properties. These properties correspond to the object's mass, as well as the static and kinetic friction coefficients for each rigid subpart of a given object.  Using a different model for each subpart  of an object is crucial to modeling articulated objects.  In this work, we focus on non-articulated objects. We divide the surface of an object  into a regular grid and identify the friction parameters  of each  part of the grid.
These properties are represented as a $d$-dimensional vector $\theta$.  A prior distribution $P_0$ on $\theta$ is used instead of a single value of $\theta$, since different instances of the same category usually have different mechanical properties. 

The online learning algorithm takes as input a prior distribution $P_t$ on the model parameters $\theta$. $P_t$ is calculated based on an initial distribution $P_0$ and a sequence of observations $(x_0,\mu_0, x_1,\mu_1, \dots, x_{t-1},\mu_{t-1}, x_{t})$, wherein $x_t$ is the 6D pose (position and orientation) of the manipulated object at time $t$ and $\mu_{t}$ is a vector describing a force applied by the robot's fingertip on the object at time $t$. Applying a force $\mu_t$ results in changing the object's pose from $x_{t}$ to $x_{t+1}$.  

\subsection{Model Identification}

Given a prior distribution $P_t$ and a new observation $(x_{t},\mu_{t+1}, x_{t+1})$, a physics engine is used to estimate a posterior distribution $P_{t+1}$ on the model parameters $\theta$. We are currently using the {\it Bullet} physics engine~\cite{Bullet}. The posterior distribution $P_{t+1}$ is obtained by simulating the effect of force $\mu_{t+1}$ on the object under various values of parameters $\theta$ and observing the resulting positions $\hat{x}_{t+1}$.  The goal is to identify the model parameters that make the outcome $\hat{x}_{t+1}$ of the simulation as close as possible to the actual observed outcome $x_{t+1}$.  In other terms, the following black-box optimization problem is solved: 
\begin{eqnarray*}
\theta^* = \arg \min_{\theta} E(\theta)  \stackrel{def}{=}  \|x_{t+1} - f(x_{t}, \mu_t, \theta) \|_2,
\end{eqnarray*}
wherein $x_{t}$ and $x_{t+1}$ are the observed poses of the object at times $t$ and $t+1$,  $\mu_t$ is the force that moved the object from  $x_{t}$ to $x_{t+1}$, and $f(x_{t}, \mu_t, \theta) =  \hat{x}_{t+1}$, the simulated pose at time $t+1$ after applying force $\mu_t$ in pose $x_t$. 

The model parameters $\theta$ can be limited to a discrete set, i.e. $\theta \in  \{\theta^1,\theta^2,\dots, \theta^n\} \stackrel{def}{=} \Theta  $.  A naive approach of solving this problem consists of systematically simulating all the parameters $\theta^i$ in $\Theta$, simulating the effect of force $\mu_t$ on the object with parameters $\theta^i$, and comparing the predicted pose  $f(x_{t}, \mu_t, \theta^i)$ to the actual pose $x_{t+1}$. However, this would be inefficient due to the size of $\Theta$, which is relatively large given that the dimension $d$ of the parameter space is typically high. Furthermore, each individual simulation is also computationally expensive. It is therefore important to minimize the number of simulations while searching for the optimal parameters. Moreover, the optimization problem above is ill-posed, as is the case in all inverse problems. In other terms, there are multiple model parameters that can explain an observed movement of an object. Instead of returning a single answer, the proposed algorithm returns a  distribution $P_{t+1}$ on the set of possible parameters $\Theta$.
 
This paper formulates this challenge in a Bayesian optimization framework, which uses the Entropy Search technique presented in~\cite{HennigSchuler2012}. This work instead presents a more computationally efficient version of the Entropy Search technique, that we call {\it Greedy Entropy Search} and describe in the following.

To solve the aforementioned Bayesian optimization problem, the error function $E$ must be learned from a minimum number of simulations, using a sequence of parameters $\theta_1,\theta_t,\dots,\theta_k\in \Theta$. To choose these parameters efficiently, a belief about the actual error function is maintained. This belief is a probability measure $p(E)$ over the space of all functions $E : \mathbb{R}^d \rightarrow \mathbb{R}$. A Gaussian Process (GP) is used to represent the belief $p$, which is sequentially updated using the errors $E(\theta_i)$ computed from simulation using model parameters $\theta_i$. Readers can find more details in textbooks on how Gaussian processes are updated from data and how to get the GP belief $p$ on unknown function $E$ from data points $E(\theta_i)$~\cite{RasmussenGPM}. The belief $p$ is initialized at each time instance $t$ using prior $P_t$, which represents the model distribution from the previous time-step. 

After simulating the object's motion with different model parameters  $\theta_1,\theta_t,\dots,\theta_k$, $p$ is updated using the computed simulation errors. $p$ implicitly defines another distribution $P_{min}$ on the identity of the best model parameter $\theta^*$, which can be used to select the next simulation parameter $\theta_{k+1}$.
\begin{eqnarray*}
\begin{aligned}
P_{min}(\theta) &\stackrel {def}{=} P\big(\theta = \arg\min_{\theta^i \in \Theta} E(\theta^i)\big) \\
&=  \int_{E: \mathbb{R}^d \rightarrow \mathbb{R}}  p(E) \Pi_{\theta^i \in \Theta-\{\theta\} } H \big(E(\theta^i) - E(\theta)\big)  \mathrm{d}E,
\end{aligned}
\end{eqnarray*}
where $H$ is the Heaviside step function, i.e. $H \big(E(\theta^i) - E(\theta)\big) = 1$ if $E(\theta^i) \geq E(\theta)$ and  $H \big(E(\theta^i) - E(\theta)\big) = 0$ otherwise. 

Unlike $p(E)$, the distribution of the simulation error $E$ modeled as a Gaussian Process, the distribution $P_{min}$ does not have a closed-form expression. Therefore, {\it Monte Carlo} is used for estimating $P_{min}$ from samples of $E(\theta^i)$ for each $\theta^i\in \Theta$. 
Specifically, this process samples vectors containing the values that $E$ takes, according to the learned Gaussian process, in each model parameter in $\Theta$. $P_{min}(\theta^i)$ is estimated by counting the fraction of sampled vectors of the values of $E$ where $\theta^i$ happens to have the lowest value.

The model parameter $\theta$ is chosen such that it has the highest contribution to the current entropy of $P_{min}$, i.e. with the highest term $-P_{min}(\theta) \log \big(P_{min}(\theta)\big)$, as the next model parameter to evaluate in simulation. This method is referred to as the {\it Greedy Entropy Search} method because it aims to decrease the entropy of the belief $P_{min}$. This process is repeated until the entropy of $P_{min}$ does not change much or until the simulation's time budget is consumed. After that, $P_{min}$ is used as the new belief $P_{t+1}$ on the model parameters. This new belief can then be utilized for planning an action $\mu_{t+1}$ which will move the object to a new pose $x_{t+1}$, after which the same process is repeated all over again.

\subsection{Policy Optimization}
Given a distribution $P_t$ on the model (e.g, friction parameters and mass), and cost function $J:\tau \rightarrow \mathbb R$, 
where $\tau = (x_0,\mu_0, x_1,\mu_1, \dots, x_{H-1},\mu_{H-1}, x_{H})$ is a trajectory of predicted object poses and applied forces, the robot needs to  find a feedback control policy $\pi_{\eta}$ that returns an action $\mu_t$ in pose $x_t$ of the object.  Policy $\pi_{\eta}$ is limited to a family of predefined policies (e.g, pushing directions) and parametrized by  $\eta$ (e.g., end-effector velocity along a given pushing direction).  Since the physics engine that we are using is deterministic, the transition model used by the physics engine is defined to be a function $f$ that takes as input  an initial pose $x_0$ and a policy $\pi_{\eta}$, a model parameter $\theta$ and returns a trajectory $\tau = f(x_0, \pi_{\eta}, \theta)$. We then search for a policy parameter $\eta^*$ defined as $\eta^*=\arg\min_{\eta} J(f(x_0, \pi_{\eta},\theta))$.  

To solve this problem in real-time, only the most likely object model $\theta^* = \arg\max_{\theta\in \Theta} P_t(\theta)$ is used for finding the optimal policy parameter $\theta^*$.  The policy parameter $\eta$ can be limited to a discrete set, i.e. $\eta \in  \{\eta^1,\eta^2,\dots, \eta^n\} \stackrel{def}{=} \Pi$.  A naive approach of solving this problem consists of iterating over all the parameters $\eta^i$ in $\Pi$, simulating a trajectory $\tau_i = f(x_0, \pi_{\eta_i}, \theta^*)$ of the object using policy $\pi_{\eta_i}$, and selecting the policy parameter $\eta_i$ with the minimum cost $J(\tau_i)$. However, this would be computationally inefficient.

We therefore use the same {\it Greedy Entropy Search} method, presented in the previous section, for searching for the best policy parameter $\eta^*$ in real-time. This is achieved by noticing the analogy between model parameters $\theta$ and policy parameters $\eta$, and between the simulation error $E(\theta)$ and the cost function $J(f(x_0, \pi_{\eta}, \theta^*))$. Hence, the same technique can be used for finding $\eta^*=\arg\min_{\eta} J(f(x_0, \pi_{\eta},\theta))$ where $\theta$ is known and $\eta$ is a variable.


\section{Experiments}
In all experiments, we used {\it Blender}\cite{Blender} which utilizes the {\it Bullet}~\cite{Bullet} physics engine. PHYSIM\_6DPose\cite{PHYSIM} was used to track the object and provide the initial and final poses of the object, through a RealSense depth camera mounted on the torso of the {\it Motoman} robot. Videos of the experiments can be found here: \url{https://goo.gl/8Pi2Gu}.

\subsection{Learning Physical Properties for Motion Prediction}
\label{motion prediction}

\begin{figure}[!htbp]
\centering
\includegraphics[width=0.45\textwidth]{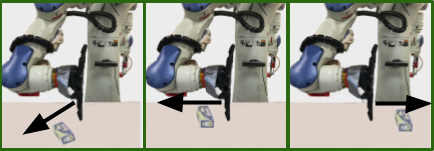}   
\caption{Data collection: the robot executes a series of random pushes, and records the location of the object before and after being pushed.}
\label{fig:moto_push_1}
\end{figure}

\subsubsection{Data Collection and Evaluation Metrics}\label{data}

In this preliminary experiment, a {\it Reflex SF} robotic hand 
mounted on the right arm of a {\it Motoman SDA10F} manipulator was used to randomly push a simple rigid object on a tabletop, as shown in Figure \ref{fig:moto_push_1}. We learn the object model parameters $\theta$ (mass and the friction coefficient) of an {\it Expo} eraser. During data collection, no human effort is needed to reset the scene since both the speed and pushing direction were controlled such that the object was always in the workspace of the robotic hand.  Using the collected pushing data, the physical properties of the object were learned so as to predict the motion of the object under new actions. Fifteen random pushing actions were performed. Six actions were discarded due to inaccurate tracking caused by occlusions. Out of the remaining nine actions, six were used for training and the other three for testing. To measure the accuracy of the learned model, the error between the predicted location of the object and the observed end location was computed.

Additionally, a large scale planar push dataset\cite{yu2016more} was also used to validate the proposed method. The dataset contained recorded poses of a planar object before and after being pushed by an ABB IRB 120 robot arm. The poses are recorded using the Vicon tracking system and are therefore more accurate.


\subsubsection{Results}
\begin{figure}[!htbp]
\begin{subfigure}[b]{0.25\textwidth}
        \centering
       \includegraphics[width=\textwidth]{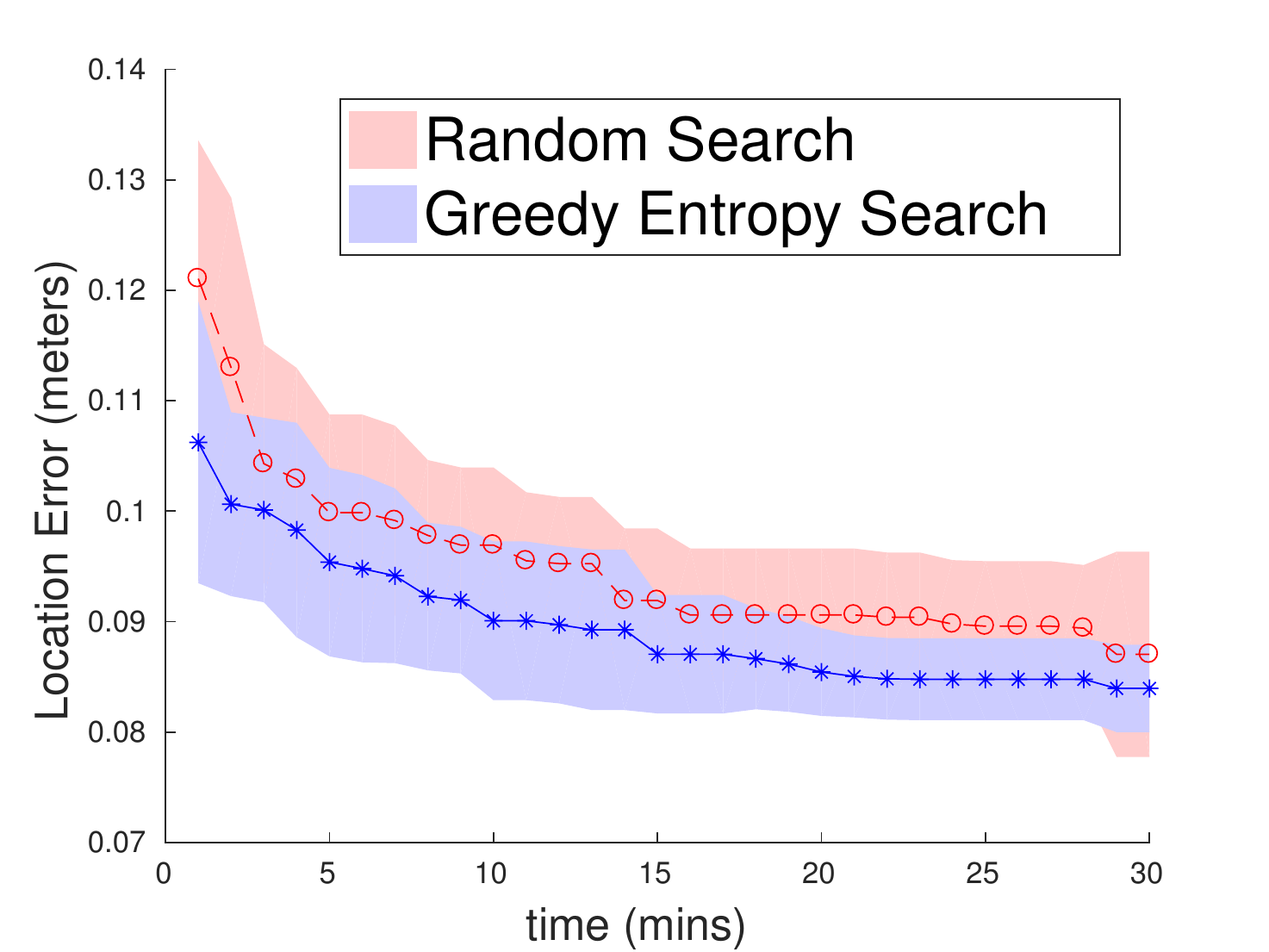}
       \caption{}
       \label{motoman_data}
\end{subfigure}%
\begin{subfigure}[b]{0.25\textwidth}
        \centering
       \includegraphics[width=\textwidth]{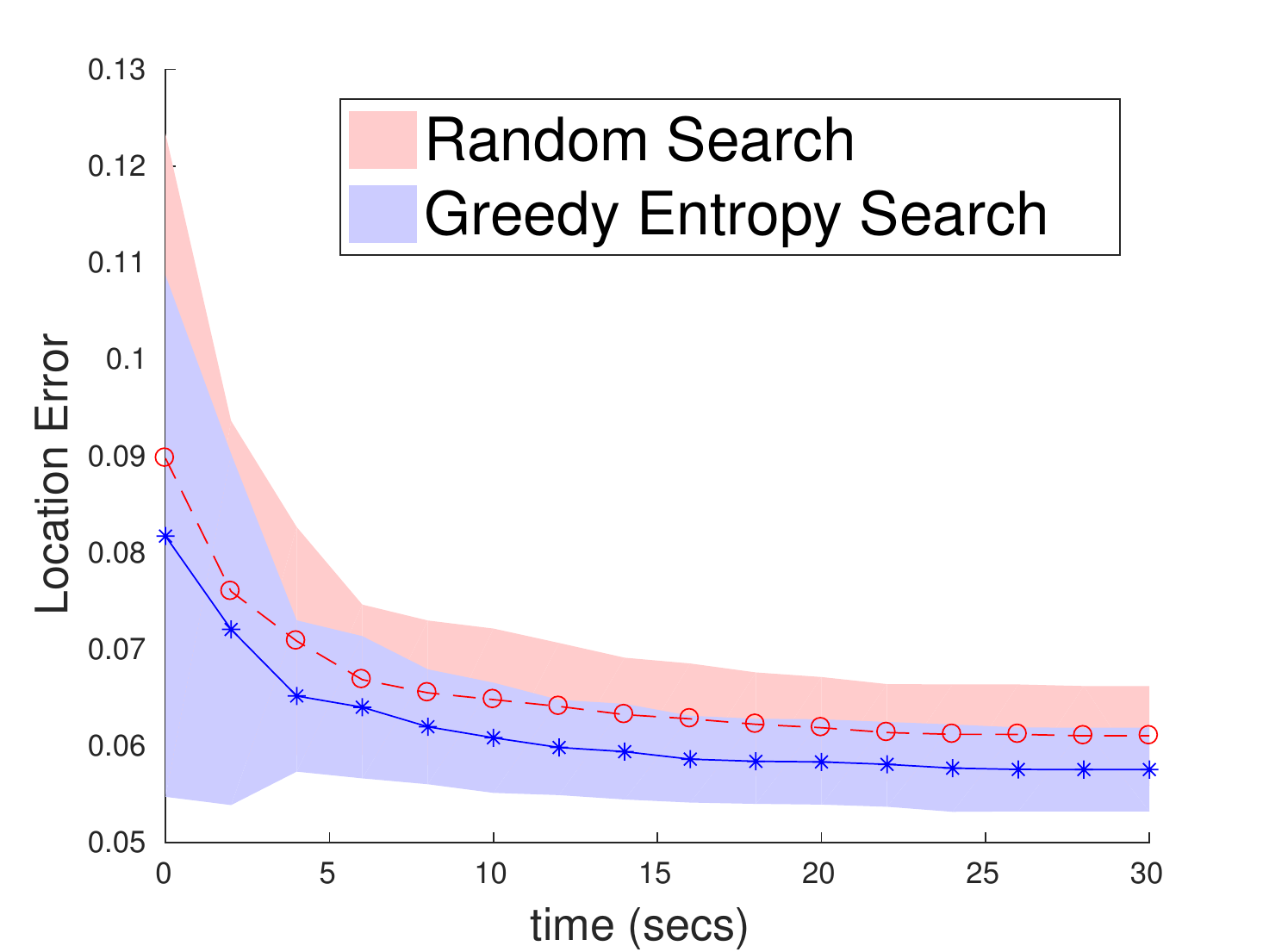}      
       \caption{}
       \label{MIT}
\end{subfigure}%

\caption{Comparison between Greedy Entropy Search method with random search. Greedy Entropy Search achieved lower error in both (a) the data collected with {\it Motoman} and (b) the planar push dataset\cite{yu2016more}.}
\label{BO_GS}
\end{figure}

We compared the results of the proposed Greedy Entropy Search method against Random Search in Figure~\ref{BO_GS}. Random Search was performed by randomly sampling $\theta$ in the same parameter space as the Greedy Entropy Search. Both methods were run ten times, with the resulting mean and standard deviation of the training error reported. The results show that Greedy Entropy Search achieved lower error when predicting the results of new actions.

\begin{figure}[!htbp]
\begin{subfigure}[b]{0.25\textwidth}
  \centering
  \includegraphics[width=\textwidth]{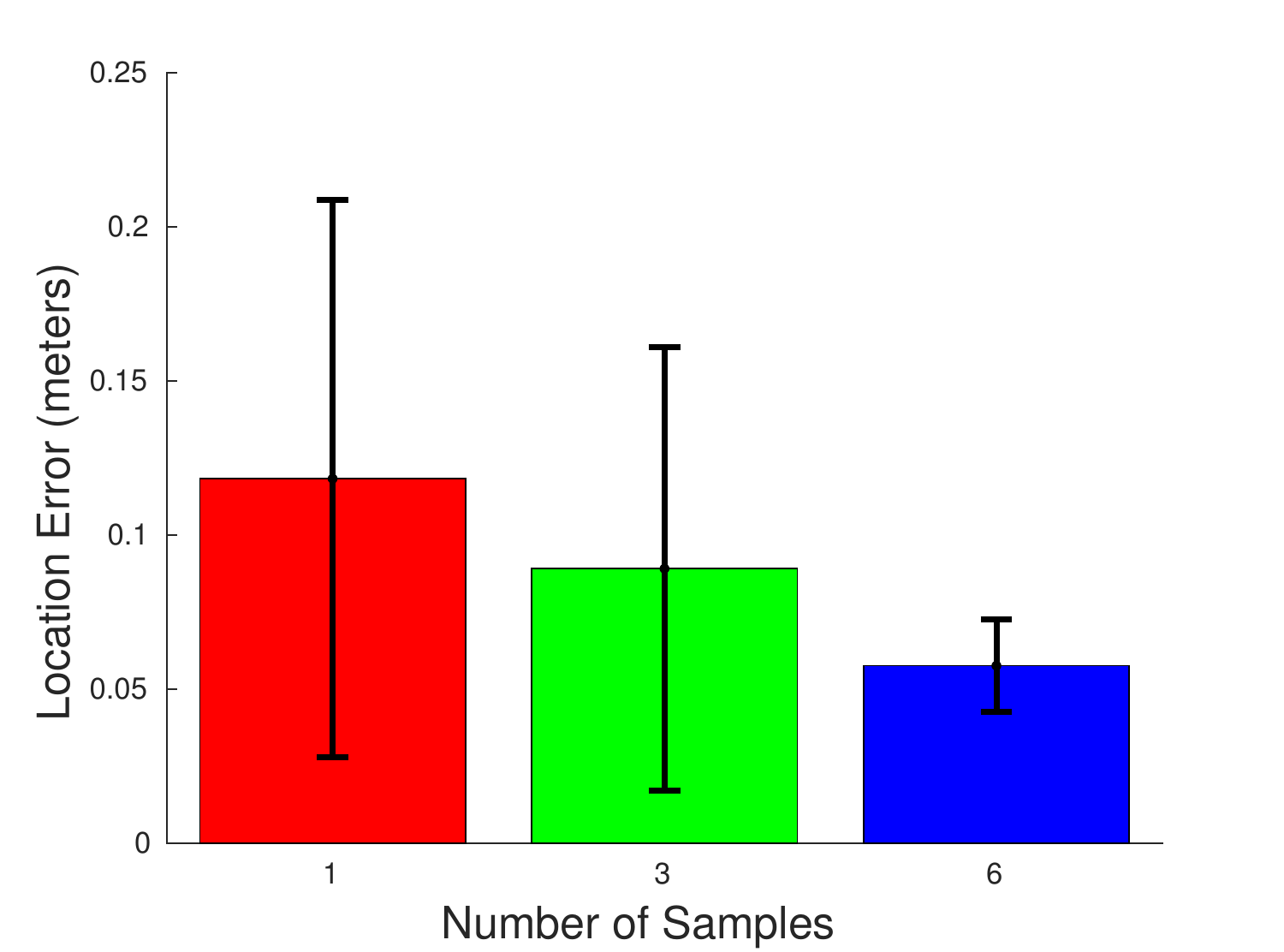}
  \caption{}
  \label{no_of_sample}
\end{subfigure}%
\begin{subfigure}[b]{0.25\textwidth}
  \centering
  \includegraphics[width=\textwidth]{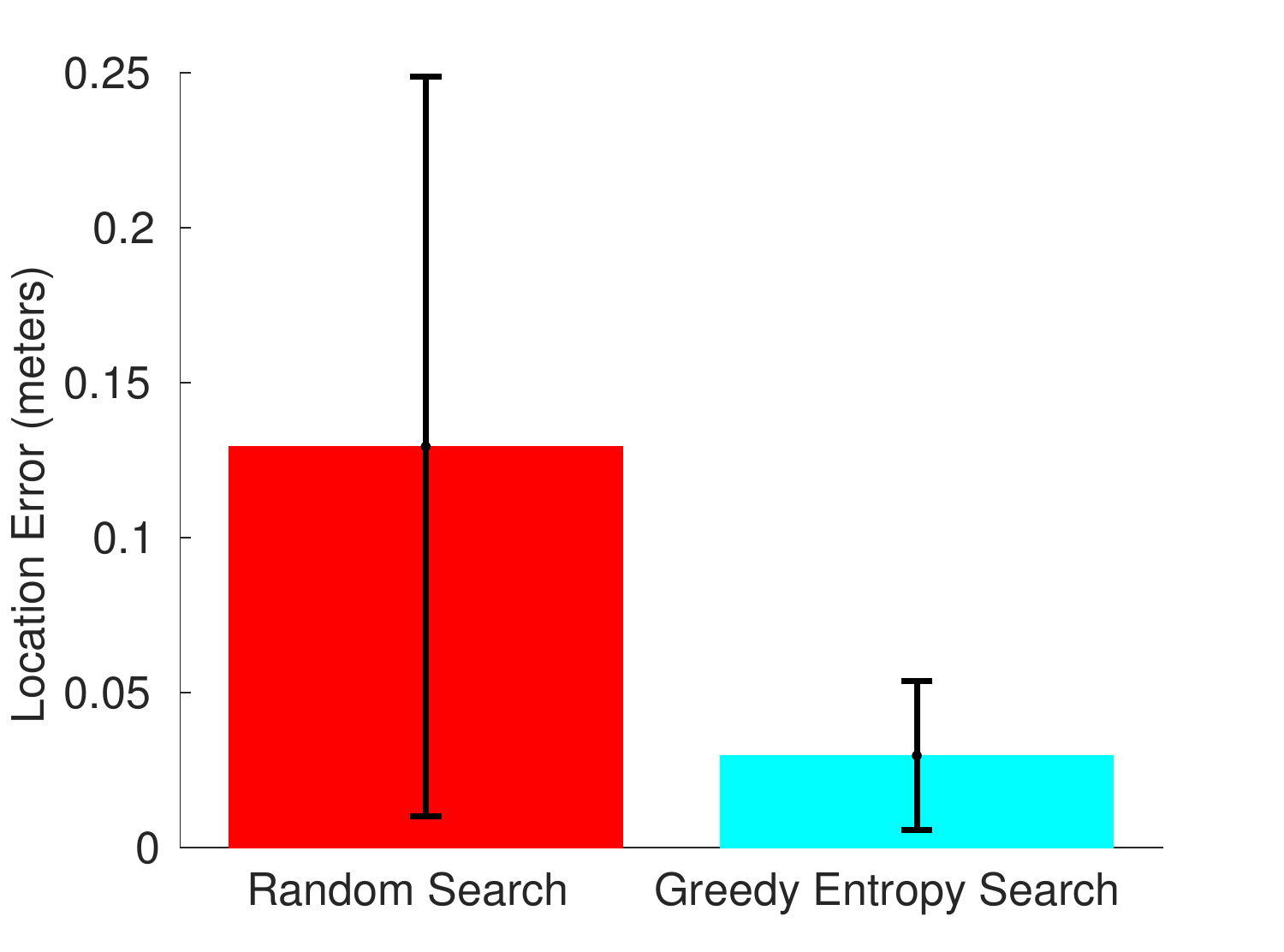}   
  \caption{}

\end{subfigure}%
  \caption{(a) Location prediction error decreases as the number of training samples increases. (b) Greedy Entropy Search achieved lower error.}

\end{figure}

The prediction error is also reported as a function of the number of training samples. Figure~\ref{no_of_sample} shows a comparison between the prediction errors of models trained with one sample, three samples and all six samples in.  The results indicate that with more training samples, the average error decreases. 


The proposed method was also tested using a large scale pushing dataset\cite{yu2016more}. Specifically, we report the result using the {\it rect1} shape on the {\it abs} surface. 200 samples were randomly selected and the result of 10-fold cross validation is shown in Figure\ref{MIT}. The proposed Greedy Entropy Search also achieved lower error than the Random Search baseline.

\subsection{Policy Optimization using the Motion Prediction Model}
\label{action_selection}

\begin{figure}[!htbp]
\centering
\includegraphics[width=0.45\textwidth]{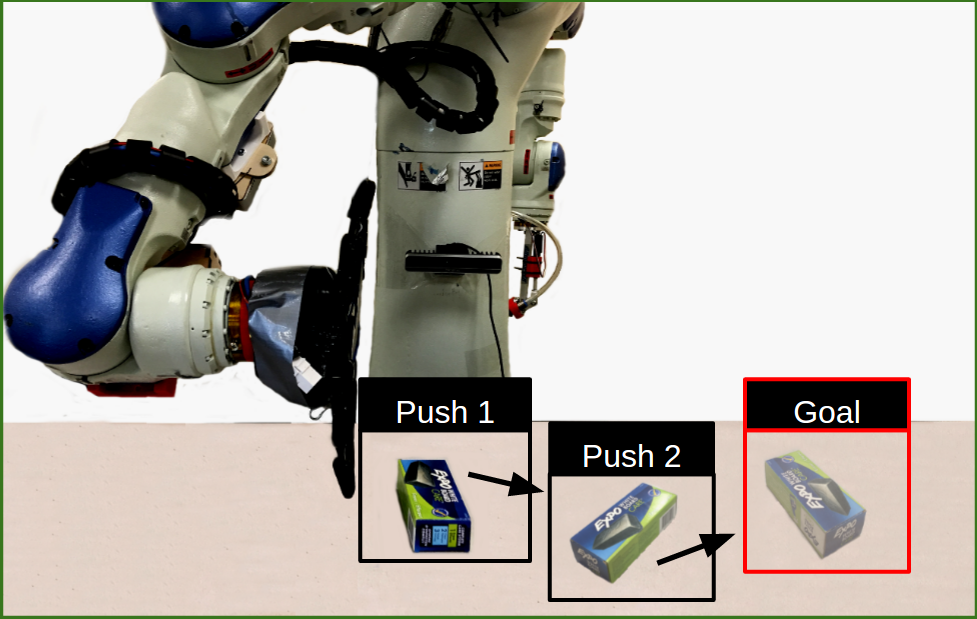}   
\caption{Once the robot has learned the physical properties of the object, it can find the optimal policy to push the object to a specific goal region.}
\label{fig:moto_push_2}
\end{figure}

\subsubsection{Setup}
In this experiment, the task is to push the object to a fixed goal position from a start region. The setup is similar to \ref{motion prediction}, a \emph{Motoman} manipulator pushing an \emph{Expo} eraser using a {\it Reflex hand}. For each trial, we push the object twice towards the goal, as shown in Figure \ref{fig:moto_push_2}. In this experiment, the policy parameter $\eta$ is the push direction. 25 random actions are sampled and the action that can push the object closest to the goal position is selected to be executed. 

\subsubsection{Results}
\begin{figure}[!htbp]
\centering
\includegraphics[width=0.5\textwidth]{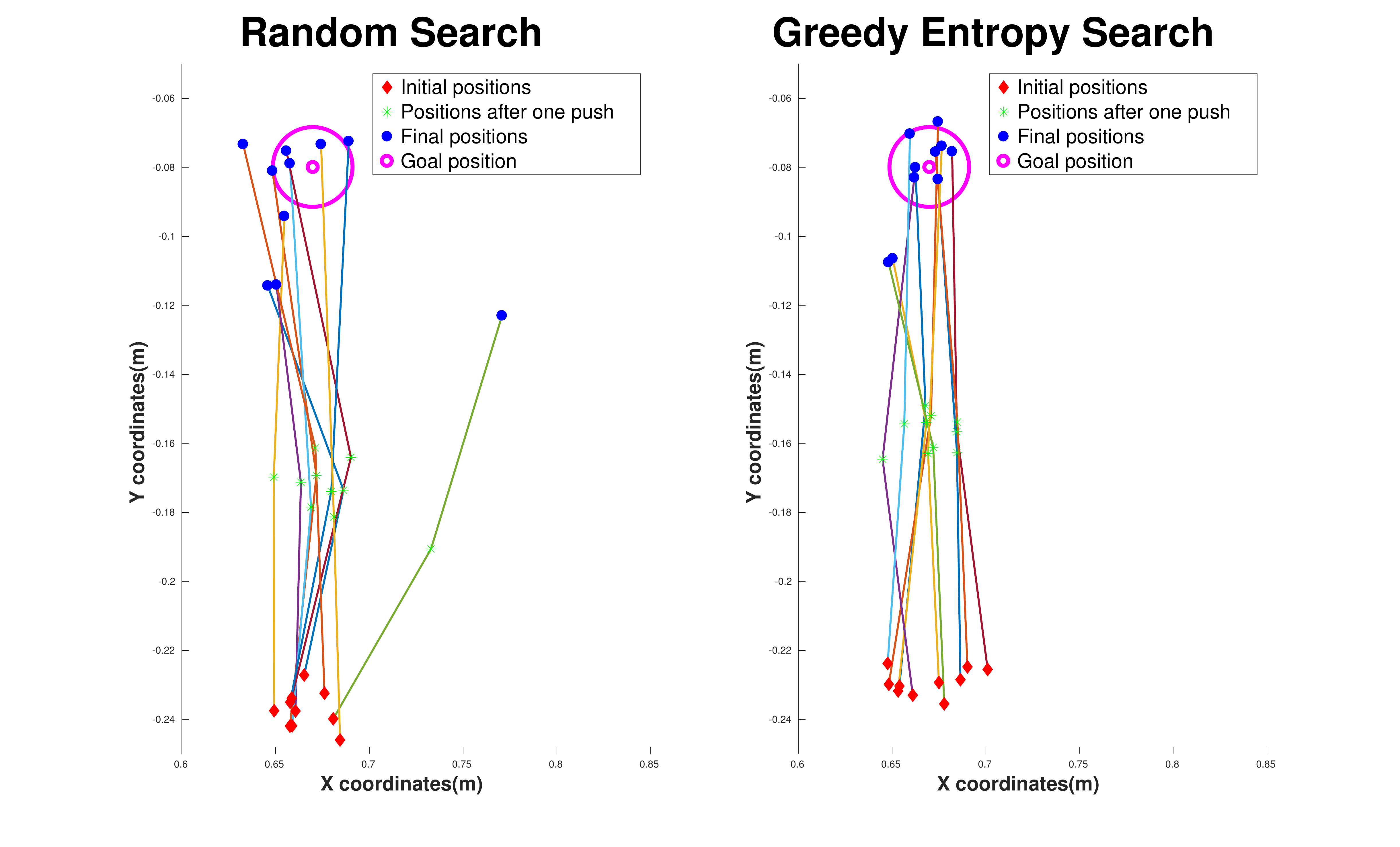}       
\caption{Comparison between Greedy Entropy Search method with Random Search for policy optimization. Greedy Entropy Search method achieves higher accuracy in pushing the object to the goal position.}
\label{action}
\end{figure}

We compare the pushing results using motion prediction model with two sets of parameters: one is learned using Greedy Entropy Search, the other is found using Random Search. Figure~\ref{action}
shows that the model using Greedy Entropy Search enabled the robot to push the object to the  1cm vicinity of the goal position 7 out of 10 trials, while the  one using Random Search only did it 4 times.

\subsection{High Speed Push Policy Optimization using Model Trained with Low Speed Push}
\label{exp:high_spped}
So far, the actions were limited to low speed pushes so that the object was always in the reachable workspace of the robot. In order to solve the challenge presented in Figure \ref{fig:ex}, however, higher speed push actions are needed. The friction between the object and the contact surface varies when the object moves at different speeds. We can collect data using higher speed push in a similar way to \ref{data}. However, this also means much more human resets will be needed, since the robot would push the object away from its workspace, sometimes even off the table. In this experiment, we avoid the human resets and aim to optimize high speed pushing policy using model trained with only low speed pushing data.

\subsubsection{Setup}

\begin{figure}[!htbp]
\centering
\includegraphics[width=0.45\textwidth]{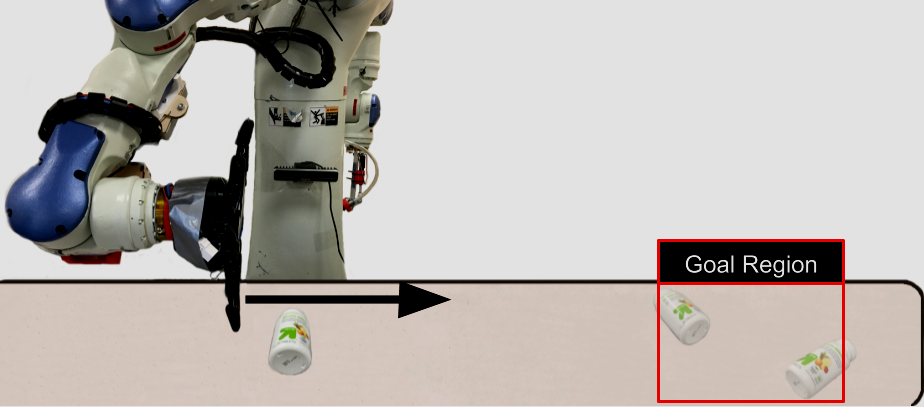}   
\caption{The task in \ref{exp:high_spped} to push the object to the other  side of the table.}
\label{fig:moto_push_3}
\end{figure}

\begin{figure*}[!tb]
\begin{center}
	       \includegraphics[width=\textwidth]{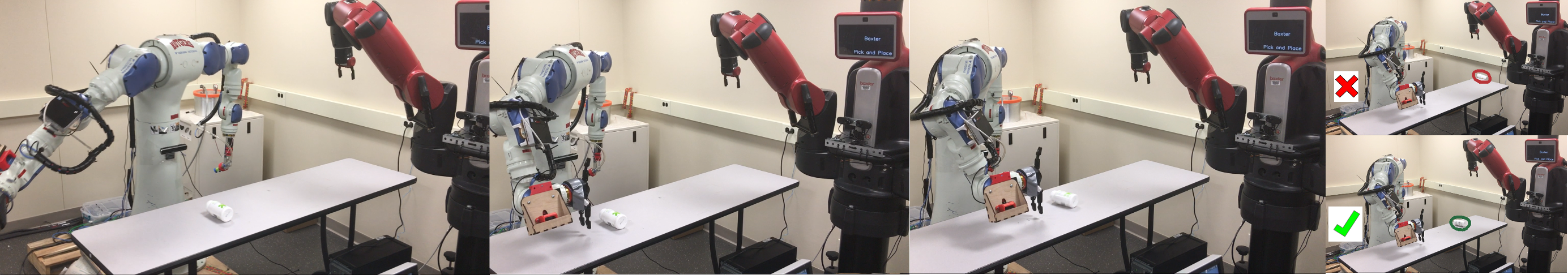}
\end{center}
\caption{Examples of experiment runs in \ref{exp:high_spped}, where the {\it Motoman} tries to push the object into Baxter's workspace without dropping it.}
\label{fig:full_push}
\end{figure*}

In this experiment, the task is to push the bottle from one side to the other side of the table, which is about one meter away, as shown in Figure \ref{fig:moto_push_3}. We aim to find the optimal policy with parameter $\eta$ representing the pushing speed of the robotic hand. We collected random low speed pushing data in a similar way to \ref{data}, using a {\it glucose bottle}, without human reset. 


After being pushed, the object sometimes is no longer within the view of the RealSense camera on the torso of {\it Motoman}. Instead, the in-hand camera on {\it Baxter} robot was used to localize the  final location of the object after it's being pushed. After learning the object model with parameters $\theta$ (mass and the friction coefficient), using the Greedy Entropy Search approach, optimal policy that can push the object closest to the goal position is selected.  We compare our approach with a model-free reinforcement learning method: Policy learning by Weighting Exploration with the Returns (PoWER)\cite{kober2009policy}. PoWER iteratively optimizes a stochastic policy as an Expectation-Maximization(EM) problem, directly using real roll-outs results.

\subsubsection{Results}

We report results from both simulation and real roll-outs. We evaluate:  
\begin{itemize}
\item The error between the final object location after being push and the desired goal location.
\item The number of times object falling off the table.
\end{itemize}

\begin{figure}[!tb]
\begin{subfigure}[b]{0.25\textwidth}
        \centering
       \includegraphics[width=\textwidth]{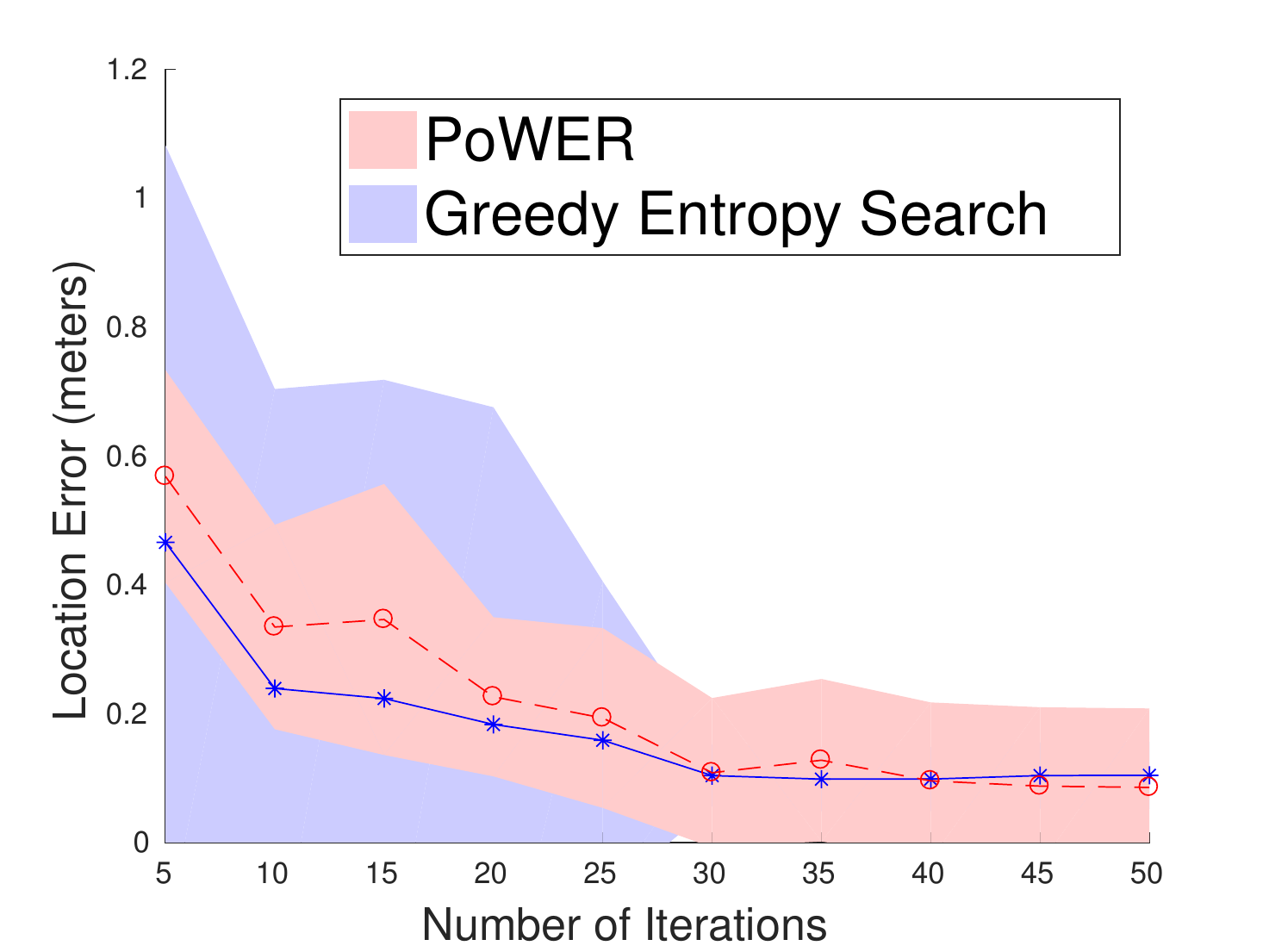}   
\end{subfigure}%
\begin{subfigure}[b]{0.25\textwidth}
        \centering
       \includegraphics[width=\textwidth]{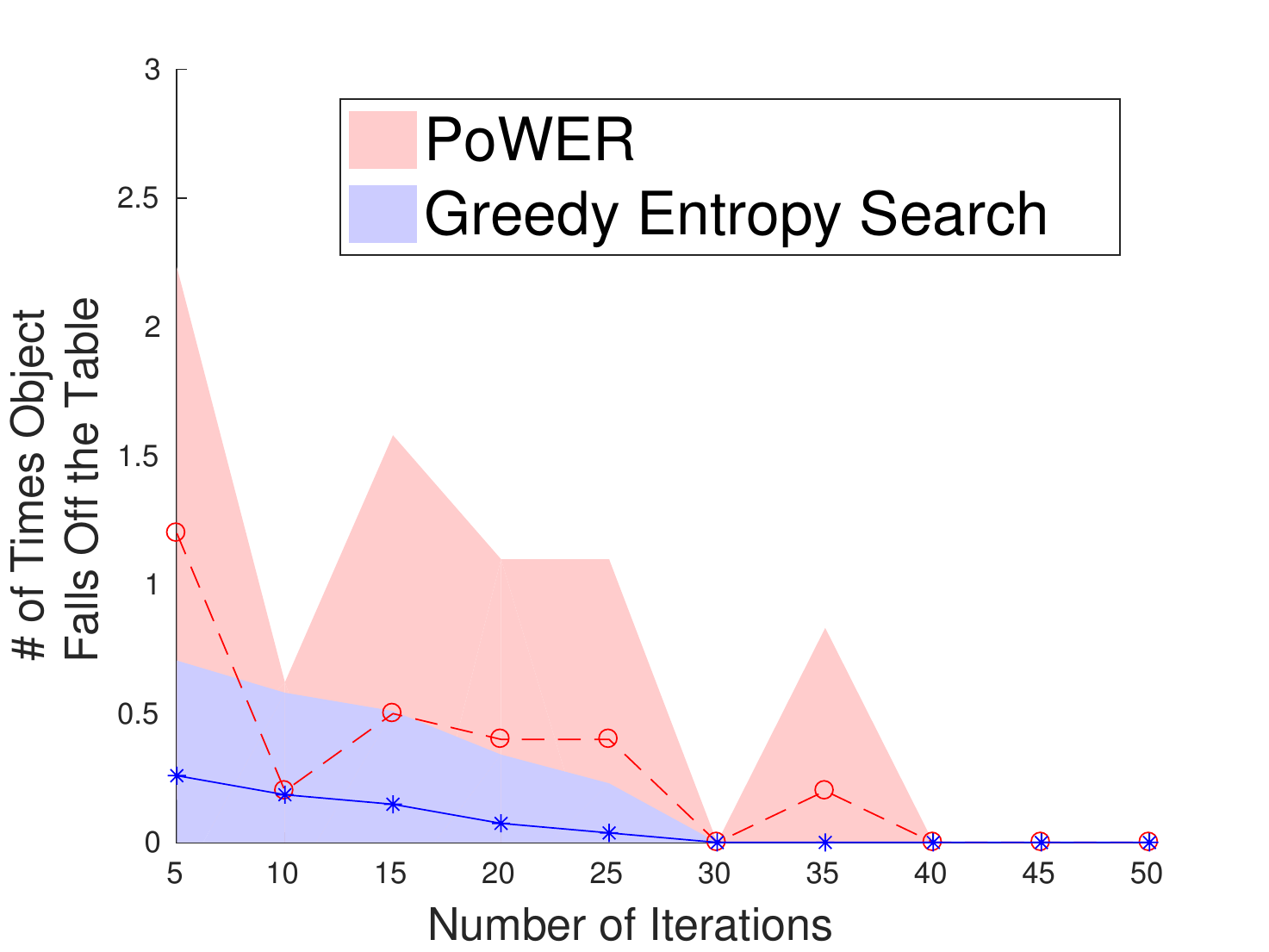}        
\end{subfigure}%
\caption{High speed push policy optimization result in simulation. Proposed Greedy Entropy Search achieves faster convergence and fewer object drops comparing with model-free reinforcement learning method PoWER.}
\label{fig:Simulation}
\end{figure}

\begin{figure}[!tb]
\begin{subfigure}[b]{0.25\textwidth}
        \centering
       \includegraphics[width=\textwidth]{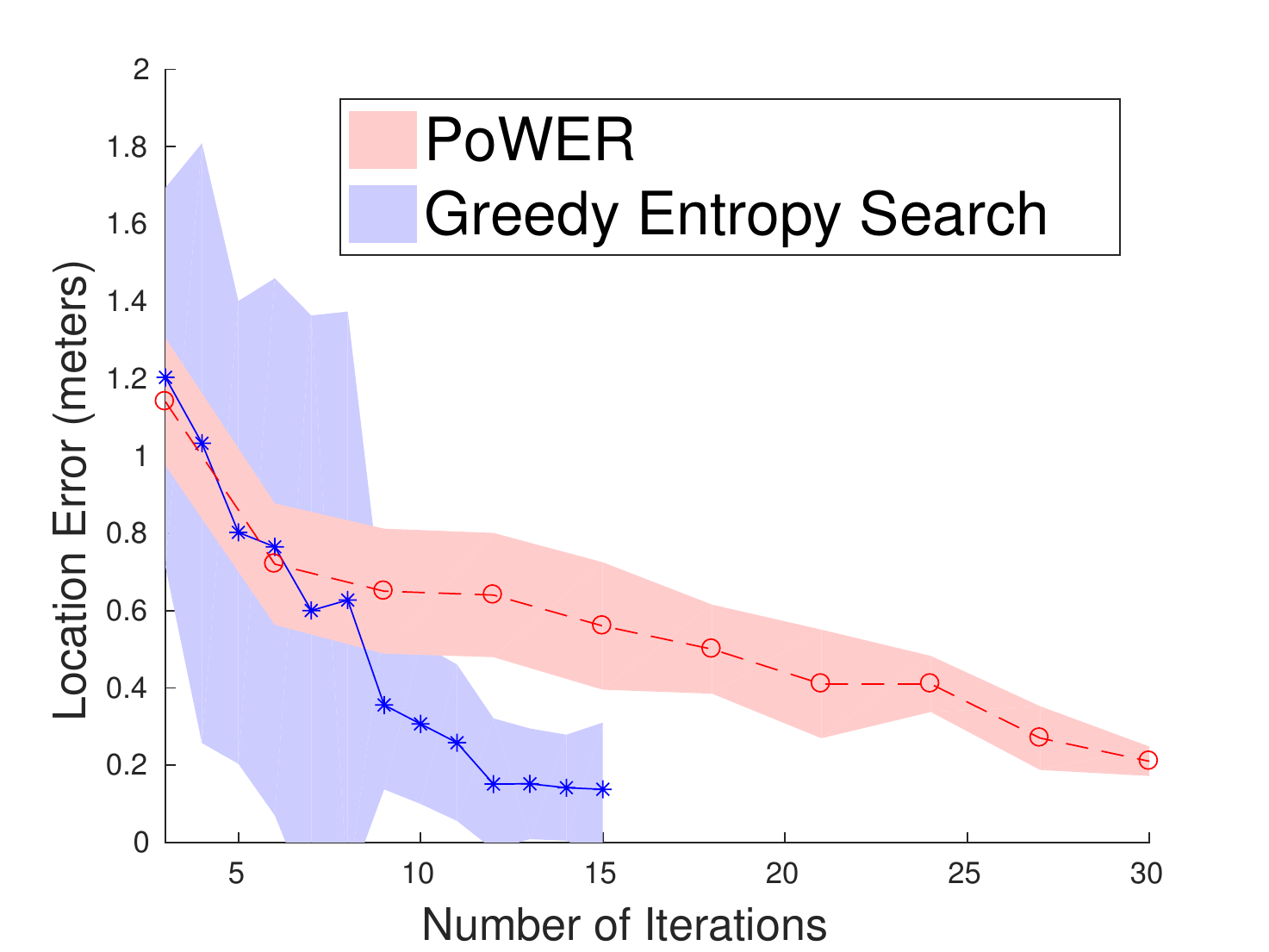}   
\end{subfigure}%
\begin{subfigure}[b]{0.25\textwidth}
        \centering
       \includegraphics[width=\textwidth]{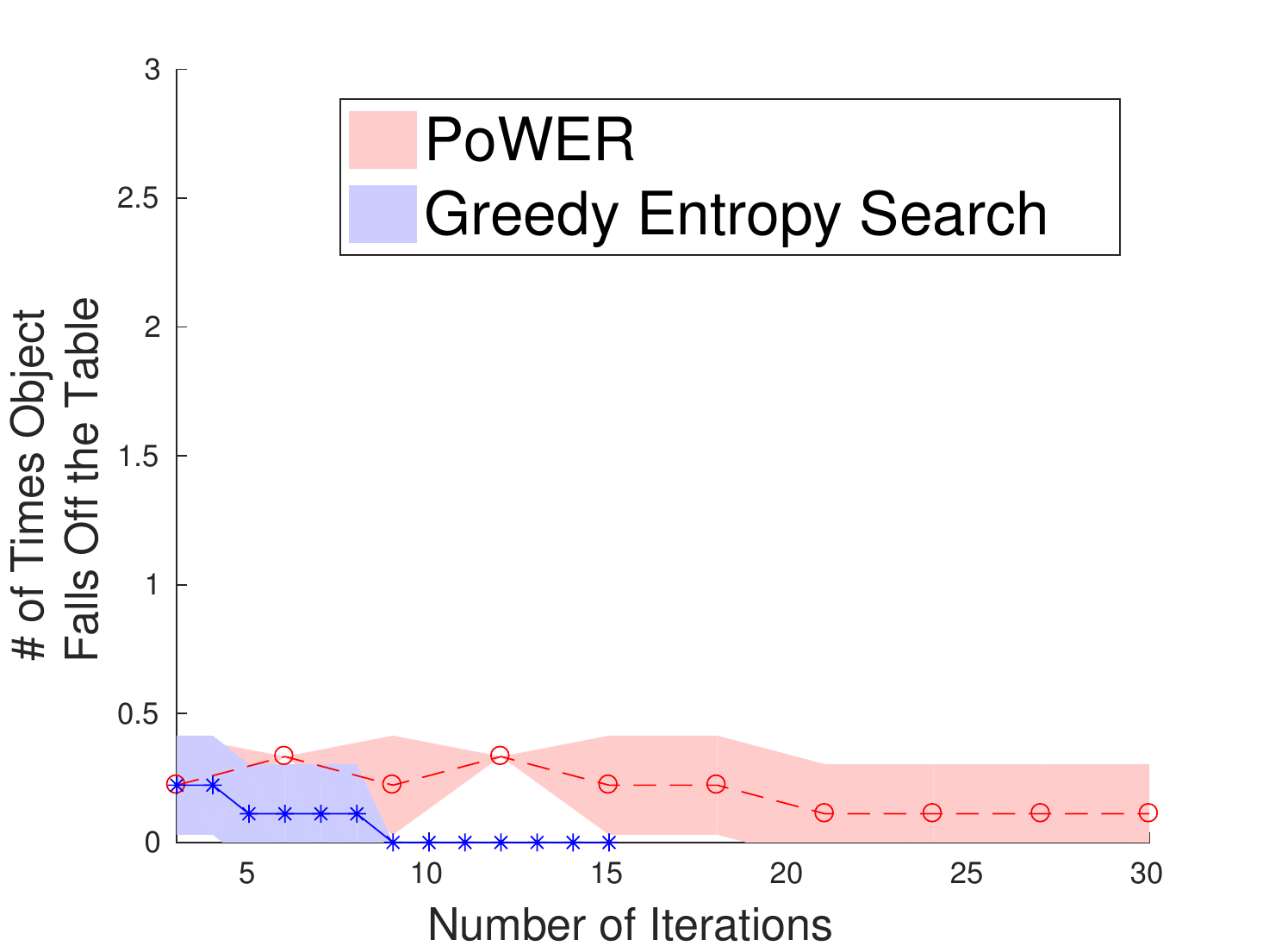}        
\end{subfigure}%
\caption{High speed push policy optimization result using a real {\it Motoman} robot.}
\label{fig:Rollouts}
\end{figure}

Figure \ref{fig:Simulation} and \ref{fig:Rollouts} show the result in simulation and with a real {\it Motoman} robot. In simulation, we randomly set ground-truth(GT) mass and friction parameters and perform roll-outs using the GT parameters. Both in simulation and with the real {\it Motoman} robot, the proposed method achieves both lower error and fewer object drops. We argue this is important in robot learning as we would like to minimize human efforts during the learning process in order to achieve autonomous robot learning. Notice that PoWER achieved smaller variance in real rollouts comparing to simulation. The probable reason for that is that because of sensing and actuation error in real roll-outs, PoWER tended to be over conservative in terms of pushing speed because of the object drops it made.

\section{Conclusion and Future Work}
In this paper, we presented a data-efficient online learning method for identifying mechanical properties of objects. The method leverages a physics engine through simulation and finds the optimal parameters that match the real roll-outs in a Bayesian optimization framework. The same framework is also used for policy optimization. Experimental results, both in simulation and using a real robot, show that the method outperforms model-free reinforcement learning methods.

An important aspect of robot learning is how many real world roll-out data are enough to achieve a certain success rate. We are currently working on evaluating the model confidence by computing the expected success rate using the uncertainty of the model. In the future, finding efficient methods for handling model parameters of non-homogeneous objects is an interesting future direction that can help scaling to more complex environment. Furthermore, while this work only considered random exploratory actions, a more intelligent way of action sampling could help better exploring the action space. Additionally, it would be interesting to investigate combining the pre-trained deep models with online learning to achieve both high capability of generalization and data efficiency.


\bibliographystyle{unsrt}
{\bibliography{references}}

\begin{thebibliography}{10}

\bibitem{agrawal2016learning}
Pulkit Agrawal, Ashvin Nair, Pieter Abbeel, Jitendra Malik, and Sergey Levine.
\newblock Learning to poke by poking: Experiential learning of intuitive
  physics.
\newblock {\em NIPS}, 2016.

\bibitem{Bullet}
{Bullet physics engine}.
\newblock [Online]. Available: \url{ www.bulletphysics.org}.

\bibitem{MuJoCo}
{MuJoCo physics engine}.
\newblock [Online]. Available: \url{www.mujoco.org}.

\bibitem{DART}
{DART physics egnine}.
\newblock [Online]. Available: \url{ http://dartsim.github.io}.

\bibitem{PhysX}
{PhysX physics engine}.
\newblock [Online]. Available: \url{www.geforce.com/hardware/technology/physx}.

\bibitem{Havok}
{Havok physics engine}.
\newblock [Online]. Available: \url{ www.havok.com}.

\bibitem{ODE}
{Open dynamics engine}.
\newblock [Online]. Available: \url{http://ode.org}.

\bibitem{GraspItSimulator}
Andrew Miller and Peter~K. Allen.
\newblock {Graspit!: A Versatile Simulator for Robotic Grasping}.
\newblock {\em IEEE Robotics and Automation Magazine}, 11:110--122, 2004.
\newblock http://www.cs.columbia.edu/~cmatei/graspit/.

\bibitem{ErezTT15}
Tom Erez, Yuval Tassa, and Emanuel Todorov.
\newblock Simulation tools for model-based robotics: Comparison of bullet,
  havok, mujoco, {ODE} and physx.
\newblock In {\em {IEEE} International Conference on Robotics and Automation,
  {ICRA} 2015, Seattle, WA, USA, 26-30 May, 2015}, pages 4397--4404, 2015.

\bibitem{Sutton:1998:IRL:551283}
Richard~S. Sutton and Andrew~G. Barto.
\newblock {\em Introduction to Reinforcement Learning}.
\newblock MIT Press, Cambridge, MA, USA, 1st edition, 1998.

\bibitem{Hamrick2016Cognitive}
J.~Hamrick, P.~W. Battaglia, T.~L. Griffiths, and J.~B. Tenenbaum.
\newblock Inferring mass in complex scenes by mental simulation.
\newblock {\em Cognition}, 157:61--76, 2016.

\bibitem{Chang2016}
M.~B. Chang, T.~Ullman, A.~Torralba, and J.~B. Tenenbaum.
\newblock A compositional object-based approach to learning physical dynamics.
\newblock {\em Under review as a conference paper for ICLR}, 2017.

\bibitem{BattagliaNIPS2016}
P.~Battaglia, R.~Pascanu, M.~Lai, D.~J. Rezende, and K.~Koray.
\newblock Interaction networks for learning about objects, relations and
  physics.
\newblock In {\em Advances in Neural Information Processing Systems}, 2016.

\bibitem{Dogar_2012_7076}
Mehmet Dogar, Kaijen Hsiao, Matei Ciocarlie, and Siddhartha Srinivasa.
\newblock {Physics-Based Grasp Planning Through Clutter}.
\newblock In {\em Robotics: Science and Systems VIII}, July 2012.

\bibitem{LunchMason1996}
K.~M. Lynch and M.~T. Mason.
\newblock Stable pushing: Mechanics, control- lability, and planning.
\newblock {\em IJRR}, 18, 1996.

\bibitem{Mericli2014}
Tekin Meriçli, Manuela Veloso, and H.Levent Akin.
\newblock {Push-manipulation of Complex Passive Mobile Objects Using
  Experimentally Acquired Motion Models}.
\newblock {\em Autonomous Robots}, pages 1--13, 2014.

\bibitem{isbell:physics:2014}
Jonathan Scholz, Martin Levihn, Charles~L. Isbell, and David Wingate.
\newblock {A Physics-Based Model Prior for Object-Oriented MDPs}.
\newblock In {\em Proceedings of the 31st International Conference on Machine
  Learning (ICML)}, pages 1089--1097, 2014.

\bibitem{ZhouPBM16}
Jiaji Zhou, Robert Paolini, J.~Andrew Bagnell, and Matthew~T. Mason.
\newblock A convex polynomial force-motion model for planar sliding:
  Identification and application.
\newblock In {\em 2016 {IEEE} International Conference on Robotics and
  Automation, {ICRA} 2016, Stockholm, Sweden, May 16-21, 2016}, pages 372--377,
  2016.

\bibitem{Deisenroth:2011fu}
M.~Deisenroth, C.~Rasmussen, and D.~Fox.
\newblock {Learning to Control a Low-Cost Manipulator using Data-Efficient
  Reinforcement Learning}.
\newblock In {\em Robotics: Science and Systems (RSS)}, 2011.

\bibitem{EslamiHWTKH16}
S.~M.~Ali Eslami, Nicolas Heess, Theophane Weber, Yuval Tassa, Koray
  Kavukcuoglu, and Geoffrey~E. Hinton.
\newblock Attend, infer, repeat: Fast scene understanding with generative
  models.
\newblock 2016.

\bibitem{FragkiadakiALM15}
Katerina Fragkiadaki, Pulkit Agrawal, Sergey Levine, and Jitendra Malik.
\newblock Learning visual predictive models of physics for playing billiards.
\newblock In {\em ICLR}, 2016.

\bibitem{ullman:cogsci2014}
Tomer~D. Ullman, Andreas Stuhlm\"{u}ller, Noah~D. Goodman, and Joshua~B.
  Tenenbaum.
\newblock Learning physics from dynamical scenes.
\newblock In {\em Proceedings of the Thirty-Sixth Annual Conference of the
  {C}ognitive {S}cience {S}ociety}, 2014.

\bibitem{WuYLFT15}
Jiajun Wu, Ilker Yildirim, Joseph~J. Lim, Bill Freeman, and Joshua~B.
  Tenenbaum.
\newblock In {\em NIPS}, 2015.

\bibitem{ByravanF16}
Arunkumar Byravan and Dieter Fox.
\newblock Se3-nets: Learning rigid body motion using deep neural networks.
\newblock {\em CoRR}, abs/1606.02378, 2016.

\bibitem{finn2016deep}
Chelsea Finn and Sergey Levine.
\newblock Deep visual foresight for planning robot motion.
\newblock {\em ICRA 2017}.

\bibitem{ZhangWZFT16}
Renqiao Zhang, Jiajun Wu, Chengkai Zhang, William~T. Freeman, and Joshua~B.
  Tenenbaum.
\newblock A comparative evaluation of approximate probabilistic simulation and
  deep neural networks as accounts of human physical scene understanding.
\newblock {\em CoRR}, abs/1605.01138, 2016.

\bibitem{li16arxiv}
Wenbin Li, Seyedmajid Azimi, Ales Leonardis, and Mario Fritz.
\newblock To fall or not to fall: A visual approach to physical stability
  prediction.
\newblock 2016.

\bibitem{LererGF16}
Adam Lerer, Sam Gross, and Rob Fergus.
\newblock Learning physical intuition of block towers by example.
\newblock In {\em Proceedings of the 33nd International Conference on Machine
  Learning, {ICML} 2016, New York City, NY, USA, June 19-24, 2016}, pages
  430--438, 2016.

\bibitem{DBLP:journals/corr/PintoGHPG16}
Lerrel Pinto, Dhiraj Gandhi, Yuanfeng Han, Yong{-}Lae Park, and Abhinav Gupta.
\newblock The curious robot: Learning visual representations via physical
  interactions.
\newblock {\em CoRR}, abs/1604.01360, 2016.

\bibitem{DBLP:journals/corr/0003LF16}
Wenbin Li, Ales Leonardis, and Mario Fritz.
\newblock Visual stability prediction and its application to manipulation.
\newblock {\em CoRR}, abs/1609.04861, 2016.

\bibitem{citeulike:14184576}
Misha Denil, Pulkit Agrawal, Tejas~D. Kulkarni, Tom Erez, Peter Battaglia, and
  Nando de~Freitas.
\newblock {Learning to Perform Physics Experiments via Deep Reinforcement
  Learning}, 2016.

\bibitem{DBLP:conf/iros/YuLR15}
Kuan{-}Ting Yu, John~J. Leonard, and Alberto Rodriguez.
\newblock Shape and pose recovery from planar pushing.
\newblock In {\em 2015 {IEEE/RSJ} International Conference on Intelligent
  Robots and Systems, {IROS} 2015, Hamburg, Germany, September 28 - October 2,
  2015}, pages 1208--1215, 2015.

\bibitem{HennigSchuler2012}
Philipp Hennig and Christian~J. Schuler.
\newblock {Entropy Search for Information-Efficient Global Optimization}.
\newblock {\em Journal of Machine Learning Research}, 13:1809--1837, 2012.

\bibitem{RasmussenGPM}
Carl~Edward Rasmussen and Christopher K.~I. Williams.
\newblock {\em {Gaussian Processes for Machine Learning}}.
\newblock The MIT Press, 2005.

\bibitem{Blender}
{Blender}.
\newblock [Online]. Available: \url{http://www.blender.org}.

\bibitem{PHYSIM}
{PHYSIM\_6DPose}.
\newblock [Online]. Available: \url{https://github.com/cmitash/PHYSIM_6DPose}.

\bibitem{yu2016more}
Kuan-Ting Yu, Maria Bauza, Nima Fazeli, and Alberto Rodriguez.
\newblock More than a million ways to be pushed: A high-fidelity experimental
  data set of planar pushing.
\newblock {\em arXiv preprint arXiv:1604.04038}, 2016.

\bibitem{kober2009policy}
Jens Kober and Jan~R Peters.
\newblock Policy search for motor primitives in robotics.
\newblock In {\em Advances in neural information processing systems}, pages
  849--856, 2009.

\end{thebibliography}

\end{document}